\documentclass{article}

\PassOptionsToPackage{compress}{natbib}

\usepackage{natbib}

\usepackage[preprint]{neurips_2023}

\usepackage[utf8]{inputenc} 
\usepackage[T1]{fontenc}    
\usepackage{hyperref}       
\usepackage{url}            
\usepackage{booktabs}       
\usepackage{amsfonts}       
\usepackage{nicefrac}       
\usepackage{microtype}      
\usepackage[dvipsnames]{xcolor}         
\usepackage{adjustbox}
\usepackage{amssymb}
\usepackage{amsthm}
\usepackage{microtype}
\usepackage{graphicx}
\usepackage{subfigure}
\usepackage{booktabs} 
\usepackage{pbox}
\usepackage{multirow}
\usepackage{multicol}
\usepackage{amssymb}        
\usepackage{amsmath}        
\usepackage{mathtools}        
\usepackage[capitalize]{cleveref}
\usepackage{floatrow}

\usepackage{algorithm}
\usepackage[noend]{algpseudocode}

\usepackage{wrapfig}
\usepackage{csquotes}
\usepackage{amssymb}
\usepackage{pifont}
\newcommand{\cmark}{\ding{51}}%

\usepackage{tikz}
\usetikzlibrary{calc,arrows.meta,positioning}

\usepackage{autonum}
\usepackage{todonotes}
\makeatletter
\newcommand*\iftodonotes{\if@todonotes@disabled\expandafter\@secondoftwo\else\expandafter\@firstoftwo\fi}  
\makeatother

\tikzset{
    every node/.style={font=\sffamily\small},
    main node/.style={thick,circle,draw,font=\sffamily\Large}
}
\usetikzlibrary{shapes,arrows}
\usetikzlibrary{positioning}

\usetikzlibrary{arrows,
    chains,
    decorations.markings,
    shadows, shapes.arrows,shapes, fit}

\tikzset{%
sum/.style      = {draw, circle, node distance = 2cm}, 
input/.style    = {coordinate}, 
output/.style   = {coordinate}, 
block/.style = { draw,
              thick,
              rectangle,
              minimum height = 2em,
              fill=white,
              align=center
},
wide block/.style = {
              block,
              minimum height = 3em,
              text width=2.5cm,
              minimum width = 8em,
},
dotted_block/.style={draw=black!35!white, line width=2pt, dash pattern=on 3pt off 4pt on 6pt off 4pt,
            inner ysep=3mm,inner xsep=3mm, rectangle, rounded corners}
}

\definecolor{mydarkblue}{rgb}{0,0.08,0.45}
\definecolor{mathematicablue}{rgb}{0.11, 0.25, 0.467}

\definecolor{myyellow}{rgb}{1.0, 0.49, 0.0}
\definecolor{myorange}{rgb}{0.95, 0.35, 0.14}

\hypersetup{colorlinks,citecolor={mydarkblue},urlcolor={mydarkblue}, linkcolor={red}}

\usepackage{amsmath,amsfonts,bm}

\Crefname{appendix}{App.}{Apps.}

\def\eqref#1{equation~\ref{#1}}

\def\1{\bm{1}}

\def\rvy{{\mathbf{y}}}

\def\va{{\bm{a}}}

\def\vf{{\bm{f}}}
\def\vg{{\bm{g}}}

\def\vs{{\bm{s}}}

\def\vu{{\bm{u}}}
\def\vv{{\bm{v}}}

\def\vx{{\bm{x}}}
\def\vy{{\bm{y}}}

\def\vzero{{\bm{0}}}

\def\vmu{{\bm{\mu}}}

\def\vomega{{\boldsymbol{\omega}}}

\def\vtheta{{\boldsymbol{\theta}}}
\def\veta{{\boldsymbol{\eta}}}

\def\evtheta{{\theta}}

\def\evf{{f}}

\def\mB{{\bm{B}}}

\def\mH{{\bm{H}}}
\def\mI{{\bm{I}}}
\def\mJ{{\bm{J}}}

\def\mQ{{\bm{Q}}}

\def\mW{{\bm{W}}}

\def\mLambda{{\bm{\Lambda}}}
\def\mSigma{{\bm{\Sigma}}}

\DeclareMathAlphabet{\mathsfit}{\encodingdefault}{\sfdefault}{m}{sl}
\SetMathAlphabet{\mathsfit}{bold}{\encodingdefault}{\sfdefault}{bx}{n}

\newcommand{\R}{\mathbb{R}}

\newcommand{\Z}{\mathbb{Z}}

\DeclareMathOperator*{\argmax}{arg\,max}
\DeclareMathOperator*{\argmin}{arg\,min}

\DeclareMathOperator{\Tr}{Tr}

\newcommand{\mr}[1]{\mathrm{#1}}
\def\T2{{\mr{T(2)}}}
\def\SO2{{\mr{SO(2)}}}
\def\SE2{{\mr{SE(2)}}}

\newcommand{\vtfc}{\mathbf{w}}
\newcommand{\vtconv}{\bar{\vtheta}}
\newcommand{\vtfca}{\vtheta}
\newcommand{\vtconva}{\bar{\vtheta}}
\newcommand{\vtffca}{\vtheta_{1}}
\newcommand{\vtffcb}{\vtheta_{2}}
\newcommand{\vsfc}{\vs^{\omega}}
\newcommand{\vsfca}{\vs_{1}^{\omega}}
\newcommand{\vsfcb}{\vs_{2}^{\omega}}
\newcommand{\vsconva}{\bar{\vs}^{\omega}}

\newcommand{\vanchor}{\vu}
\newcommand{\anchor}{u}

\title{Learning Layer-wise Equivariances \\ Automatically using Gradients}

\usepackage{authblk}

\author[1]{Tycho F.A. van der Ouderaa}
\author[2,3]{Alexander Immer}
\author[1]{Mark van der Wilk}
\affil[1]{Department of Computing, Imperial College London, United Kingdom}
\affil[2]{Department of Computer Science, ETH Zurich, Switzerland}
\affil[3]{Max Planck Institute for Intelligent Systems, T\"ubingen, Germany}

\begin{document}
\newfloatcommand{capbtabbox}{table}[][\FBwidth]

\maketitle

\vspace{-1.6em}


\begin{abstract}
Convolutions encode equivariance symmetries into neural networks leading to better generalisation performance. However, symmetries provide fixed hard constraints on the functions a network can represent, need to be specified in advance, and can not be adapted. Our goal is to allow flexible symmetry constraints that can automatically be learned from data using gradients. Learning symmetry and associated weight connectivity structures from scratch is difficult for two reasons. First, it requires efficient and flexible parameterisations of layer-wise equivariances. Secondly, symmetries act as constraints and are therefore not encouraged by training losses measuring data fit. To overcome these challenges, we improve parameterisations of soft equivariance and learn the amount of equivariance in layers by optimising the marginal likelihood, estimated using differentiable Laplace approximations. The objective balances data fit and model complexity enabling layer-wise symmetry discovery in deep networks. We demonstrate the ability to automatically learn layer-wise equivariances on image classification tasks, achieving equivalent or improved performance over baselines with hard-coded symmetry.
\end{abstract}

\let\thefootnote\relax\footnotetext{\vspace{-2em}Code accompanying this paper is available at \url{https://github.com/tychovdo/ella}}

\vspace{-1.0em}
\section{Introduction}
Symmetry constraints, such as layer-wise equivariances of convolutional layers, allow neural networks to generalise beyond training data and achieve high test performance \citep{cohen2016group}. However, it is unclear which symmetries to use and to what degree they should be enforced. In digit recognition, some robustness to rotation is desirable, but full rotational invariance might prevent distinguishing 6's and 9's. Similarly, translation equivariance prevents the rate coding of positional information \citep{sabour2017dynamic}, which may be relevant for some tasks. We propose a method to automatically learn the type and amount of layer-wise equivariance symmetries from training data. 

To do so, we relax symmetries building upon recent literature on approximate equivariance, including residual pathways \citep{finzi2021residual} and non-stationary filters \citep{van2022relaxing}, effectively allowing controllable interpolation between strict equivariance or less constrained functions. To remain practical, we propose extensions to keep the parameter count similar to classic convolutions.

Allowing for differentiable selection between equivariances alone is not enough, as symmetries act as constraints and are therefore not encouraged by training losses that measure data fit, as noted in \citet{van2021learning,immer2021scalable}. Search over many continuous hyperparameters with cross-validation is infeasible, relying on expensive retraining, hold-out validation data and does not leverage gradients. Instead, we follow Bayesian model selection through differentiable Laplace approximations to learn layer-wise equivariances from training data in a single training procedure.

We demonstrate automatically learning layer-wise symmetry structure on image classification tasks. To do so, we improve upon existing parameterisations of differentiable equivariance and derive corresponding Kronecker-factored Laplace approximations to the marginal likelihood. On image classification, we show that our method automatically learns convolutional structure in early layers and achieves similar or improved performance compared to architectures with hard-coded symmetry.

\newpage
\section{Related Work}

\paragraph{Enforcing strict equivariance into the architecture.}

It can be shown that enforcing equivariance constraints in linear neural network layers is equivalent to performing a convolution \citep{kondor2018generalization,cohen2019general}. Regular convolutional layers embed equivariance to translation, but this notion can be extended to other groups by convolving over groups \citep{cohen2016group}. Group convolutions have been proposed for various domains and more complex continuous groups \citep{weiler20183d,weiler2019general,worrall2017harmonic}. Equivariances enforce hard constraints on functions and can even form a linear subspace that can be explicitly computed \citep{van2020mdp,finzi2021practical}.

\paragraph{Place coding, rate coding, disentanglement and part-whole hierarchies.}

Equivariance closely relates to disentangled representations between place-coded and rate-coded features found in capsule networks \citep{sabour2017dynamic,hinton2018matrix,kosiorek2019stacked}, as formalised in \cite{cohen2014learning,cohen2018intertwiners}. Positional information under strict translation equivariance of convolutions can only be place-coded, not rate-coded in features. Capsule networks aim to break this strict disentanglement to allow the embedding of part-whole hierarchies \citep{sabour2017dynamic}. We argue that our method solves the same problem by relaxing the disentanglement of strict equivariance and provide an automatic procedure to learn the extent to which information is exchanged and moved from place-coded to rate-coded features.

\vspace{-0.8em}
\paragraph{Approximate equivariance.}

Symmetries are often assumed known, must be fixed, and can not be adapted. This can be desirable if symmetries are known \citep{veeling2018rotation}, but becomes overly restrictive in other real-world tasks where symmetries or the appropriate amount of symmetry is not clear. Enforcing symmetry regardless can hamper performance \citep{liu2018intriguing,wang2022approximately}.

Approximate symmetry can prevent symmetry misspecification without losing the benefits that inductive biases of symmetries provide. Although strict group symmetries always apply to the entire group, by the group theoretical axiom of closure, approximate symmetry can be thought of as a form of robustness towards group actions that is only locally enforced (around training data). We use relaxations of symmetry that follow notions of approximate symmetry in literature \citep{van2018learning,wang2020incorporating}. Although invariances can often easily be relaxed by simply averaging data augmentations \citep{benton2020learning,schwobel2021last,van2021learning}, parameterising approximate equivariance \citep{wang2022approximately} can be more difficult. We build upon recent attempts, including partial equivariance \citep{romero2021learning}, residual pathways \citep{finzi2021residual}, and non-stationary filters \citep{van2022relaxing,van2023sparse}. Existing relaxations of equivariance often introduce many additional parameters due to added linear layers \citep{finzi2021residual} or use of hyper networks \citep{romero2021learning,van2022relaxing}, making adoption in practice difficult. We strive to keep parameter counts close to classical convolutional layers, sometimes requiring further simplifying assumptions.

\paragraph{Symmetry discovery and objectives}

To learn symmetry, we relax layer-wise symmetries and thereby allow them to be differentiably learned with gradients, with strict symmetry as limiting case. Since adding symmetries does not directly improve training losses that rely on training fit, there is no direct encouragement of the objective function to use symmetries. To overcome this, some works have considered using validation data \citep{maile2022architectural,zhou2020meta}. Alternatively,  \cite{benton2020learning,finzi2021residual,yang2023generative,van2022relaxing} consider learning symmetry from training data only, but to do so require explicit regularisation that needs tuning. Some problems with this approach have been noted in \cite{immer2022invariance}, which, like \citet{van2018learning}, proposes to use Bayesian model selection to learn invariances in a single training procedure, use differentiable Laplace approximations to remain tractable in deep neural networks. The method only considers invariances, which can be less efficient because it relies on learning parameters in the sampler instead of learning regularized symmetry in intermediary representations. We extend this line of work to enable automatic learning of layer-wise equivariances from training data.

\paragraph{Neural architecture search}

As layer-wise equivariances imply a convolutional architectural structure \citep{cohen2019general}, symmetry discovery can be viewed as a form of neural architecture search (NAS). Our approach uses relaxed symmetries to allow differentiable search over architectural structures, similar to DARTS \citep{liu2018darts}. Instead of using validation data, however, our method uses Bayesian model selection to learn architecture from training data only.

We distinguish ourselves from large parts of AutoML literature in that we do not require expensive outer loops \citep{zoph2018learning}, external agents \citep{cubuk2018autoaugment}, or even validation data \citep{lorraine2020,yeh2022equivariance}. Our method can be interpreted as a one-shot architecture search using Laplace approximations, similar to \citep{zhou2019bayesnas}. We demonstrate equivariance learning, but see no theoretical reason why the proposed objective should be limited to symmetry discovery, making it potentially broadly applicable in architecture search. For instance, in approaches that aim to learn network depth \citep{antoran2020depth}, filter sizes \citep{romero2023dnarch} and other layer types from training data. Although interesting future work, we focus on learning layer-wise equivariances.

\paragraph{Bayesian Model Selection}

We propose to treat layer-wise equivariances as hyperparameters and learn them with Bayesian model selection -- a well-understood statistical procedure. Traditionally, model selection has only been available for models with tractable marginal likelihoods, such as Gaussian Processes \citep{rasmussen2003gaussian}, but approximations for deep learning have also been developed, such as Variational Inference schemes \citep{ober2021global}, deep Gaussian Processes \citep{dutordoir2020bayesian}, and scalable Laplace approximations \citep{immer2021scalable}, which we use. Although application in deep learning involves additional approximations, the underlying mechanism is similar to classic Automatic Relevance Determination \citep{mackay1994bayesian} to optimise exponential basis function lengthscales. Even more so, as we cast layer-wise equivariance learning as a lengthscale selection problem in \cref{sec:defining-prior}. Learning invariances with Bayesian model selection in GPs was demonstrated in \cite{van2018learning}, and extensions to neural networks exist \citep{schwobel2021last,van2021learning}, but often use lower bounds that only work on shallow networks. For deep networks, \cite{immer2022invariance,immer2023stochastic} and \cite{mlodozeniec2023hyperparameter} recently demonstrated invariance learning with marginal likelihood estimates in large ResNets.


This work describes a method that enables layer-wise learning of equivariance symmetries from training data. We formulate different priors consisting of hyperparameters that place a different mass on functions that obey equivariance symmetry constraints and functions that do not. The amount of symmetry becomes a hyperparameter we empirically learn from data using Bayesian model selection.

In \cref{sec:parameterisations}, we discuss how to efficiently parameterise relaxed equivariance in convolutional architectures. We propose additional factorisations and sparsifications in \cref{sec:factor-sparsify} to bring the parameter count of relaxed equivariance closer to that of classical convolutions. In \cref{sec:defining-prior}, we describe how the parameterisations can be used to specify the amount of equivariance in the prior.

In \cref{sec:bayes-equiv}, we deal with the objective function and discuss approximate marginal likelihood for layer-wise relaxed equivariance. We rely on recent advancements in scalable linearised Laplace approximations that use KFAC approximations of the Hessian, and in \cref{sec:kfac-factored,sec:kfac-sparsified} derive KFAC for proposed relaxed equivariance layers to use them in conjunction with this objective.

\section{Differentiable parameterisation of layer-wise equivariance symmetry}
\label{sec:parameterisations}

To learn symmetries from training data, we use the strategy of relaxing equivariance constraints through parameterisations allowing interpolation between general linear mappings and linear mappings that are constrained to be strictly equivariant. In our choice of relaxation, we strive to keep parameter counts as close as possible to classical convolutional layers to remain useful in practice. We describe parameterisations of linear, equivariant and relaxed equivariant layers for discrete 2-dimensional translation symmetry $\Z^2$, as this is the most widely used symmetry in deep learning. All layers in this work can be naturally extended to other symmetry groups, as described in \cref{sec:other-groups}.

\paragraph{Fully-connected layer} A fully-connected layer forms a linear map between an input feature map $\mathbf{x} : [0, C] \times \Z^2 \to \R$ (spatially supported at intervals $[0, X] \times [0, Y]$), with $C$ input channels outputting a $C'$ channel feature map $\mathbf{y} : [0, C'] \times \Z^2 \to \R$ with the same spatial support:
\begin{align}
\label{eq:linear-layer}
\mathbf{y}(c', x', y') = \sum_c \sum_{x, y} \mathbf{x}(c, x, y) \vtfc(c', c, x', y', x, y) 
\end{align}
where non-zero weights in $\vtfc : \Z^6 \mapsto \R$ can be represented in $C'CX^2Y^2$ parameters. This may be more clear if, instead of using real-valued functions to describe features and weights, we equivalently flatten $\mathbf{x}, \mathbf{y}$ as vectors $\vy = \mW \vx$ and write $\vtfc$ as a matrix $\mW \in \R^{C'XY\times CXY}$. Yet, we stick to the form of \cref{eq:linear-layer} as this simplifies future factorisations and generalisations. Furthermore, the notation emphasises that other layers, such as convolutions, form special cases of a fully-connected layer.

\paragraph{Convolutional layer}
We might want to constrain a layer to be \textit{strict equivariant}, so that translations in the input $(dx, dy) \in \Z^2$ yield equivalent translations in the output, mathematically:
\begin{align}
\label{eq:equivariance-condition}
\mathbf{y}(c', x' + dx, y' + dy) = \sum_c \sum_{x, y} \mathbf{x}(c, x + dx, y + dy) \vtfc(c', c, x', y', x, y) \hspace{1em} \forall dx, dy \in \Z
\end{align}
It can be shown that equivariance is equivalent (if and only if) to having a weight matrix that is \textit{stationary} along the spatial dimension, meaning it solely depends on relative coordinates $\bar{x} = x' - x$ and $\bar{y} = y' - y$ and can thus be written in terms of stationary weights $\vtconv : (c, c', \bar{x}, \bar{y}) \mapsto \R$:
\begin{align}
\label{eq:conv}
\mathbf{y}(c', x', y') = \sum_c \sum_{x, y} \mathbf{x}(c, x, y) \vtconv(c', c, x' - x, y' - y),
\end{align}
with $C'CXY$ parameters in $\vtconv$. Note that \cref{eq:conv} is exactly a convolutional layer (`convolution is all you need', Theorem 3.1 of \cite{cohen2019general}). Constraining the stationary weights $\vtconv$ to be only locally supported on a rectangular domain $\bar{x}, \bar{y} \in [-\frac{S-1}{2}, \frac{S-1}{2}]$, and 0 otherwise, gives rise to small (e.g. $3 \times 3$) convolutional filters and further reduces the required parameters to $C'CS^2$.

\subsection{Factoring and sparsifying relaxed equivariance}
\label{sec:factor-sparsify}
\vspace{-0.5em}

To learn layer-wise equivariances, we take the approach of relaxing symmetry constraints in neural network layers allowing explicit differentiable interpolation between equivariant and non-equivariant solutions. As a starting point, we consider the residual pathways of \citep{finzi2021residual}, reinterpreted as factorisation of the linear map as $\vtfc = \vtfca + \vtconva$ with fully-connected $\vtfca : (c', c, x', y', x, y) \mapsto \R$ and stationary $\vtconva(c', c, \bar{x}, \bar{y}) \mapsto \R$ with $\bar{x} {=} x' {-} x$ and $\bar{y} {=} y' {-} y$:
\begin{align}
\mathbf{y}(c', x', y')
= \underbrace{\sum_c \sum_{x, y} \mathbf{x}(c, x, y) \vtfca(c', c, x', y', x, y)}_{\text{fully-connected (FC)}} + \underbrace{\sum_c \sum_{x, y} \mathbf{x}(c, x, y) \vtconva(c', c, x' - x, y' - y)}_{\text{convolutional layer (CONV)}}
\vspace{-0.5em}
\raisetag{1.8\baselineskip}
\label{eq:rpp}
\end{align}
The layer consists of simply adding FC and CONV layer outputs. The CONV path spans the linear subspace of FC that obeys strict equivariance of \cref{eq:equivariance-condition}. As such, the combined layer is equally expressive as a linear map (we can construct a bijection between $\vtfc$ and $\vtfca + \vtconva$, as is common for relaxed equivariance \citep{van2022relaxing}). In \cite{finzi2021residual}, independent priors are placed on both sets of weights with relative variances controlling how much equivariance is encouraged. We also consider this prior, but assume it is unknown and use Bayesian model selection to automatically learn the variances that control symmetry constraints from data. Further, naive residual pathways of \cref{eq:rpp} require polynomially many parameters $C'CX^4 {+} C'CX^2$, or $C'CX^4 {+} C'CS^2$ for $S {\times} S$ filters, using square features $X{=}Y$ for simplicity. This makes using residual pathways in common architectures infeasible (see line 3 in \cref{tab:cifar10-results}). We overcome this by proposing factorisations and sparsifications bringing parameter counts of relaxed equivariance closer to convolutional layers.

\paragraph{Factored layers}
Inspired by common factorisations in convolutional architectures, such as depth-wise \citep{chollet2017xception} and group separability \citep{knigge2022exploiting}, we propose to factor fully-connected layers between input and output dimensions $\vtfca(c', c, x', y', x, y) = \vtffca(c', c, x', y')\vtffcb(c', c, x, y)$. This greatly reduces the parameters, while fully maintaining the benefits of fully-connected structure from full spatial $H{\times}W$ input and output $H'{\times}W'$ to both input and outputs channels $C{\times}C'$.
\begin{align}
\label{eq:f-fc}
\sum_c \sum_{x, y} \mathbf{x}(c, x, y) \vtfca(c', c, x', y', x, y)
&= \underbrace{\sum_c \vtffca(c', c, x', y') \sum_{x, y}\mathbf{x}(c, x, y) \vtffcb(c', c, x, y)}_{\text{factored fully-connected (F-FC)}}
\vspace{-1em}
\end{align}
\begin{wrapfigure}{r}{0.40\textwidth}
  \begin{center}
    \includegraphics[width=1.0\textwidth]{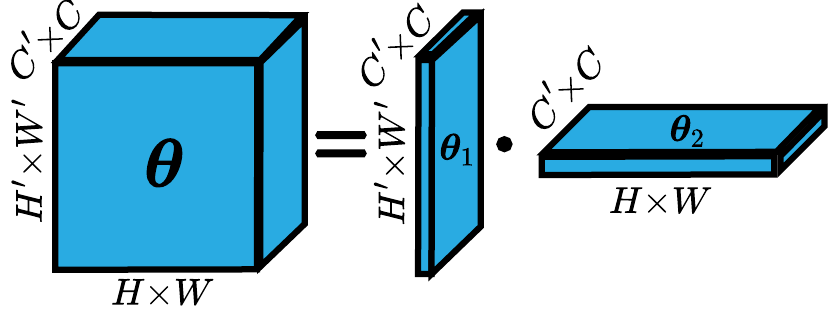}
  \end{center}
   \caption{\cref{eq:f-fc} visualised as tensors.}
\end{wrapfigure}
The factorisation of \cref{eq:f-fc} has several benefits. First, it reduces the number of parameters required for the expensive the fully-connected path from $C'CX^2Y^2$ to $2C'CXY$. This is essential in practice to obtain parameter counts close to convolutional layers (133.6M${\rightarrow}$1.5M in \cref{tab:cifar10-results}). Secondly, we can write the layer as a composition of two linear operations summing over $\sum_{x, y}$ and $\sum_c$, which we use to derive Kronecker-factored approximations to the second-order curvature and enable the use of scalable Laplace approximations. Lastly, the parameters $\vtffca$ and $\vtffcb$ have a single spatial domain that lends themselves to further optional sparsification with basis functions.

\begin{wrapfigure}{R}{0.5\textwidth}
\vspace{-7.0mm}
  \begin{center}
    \includegraphics[width=1.0\textwidth]{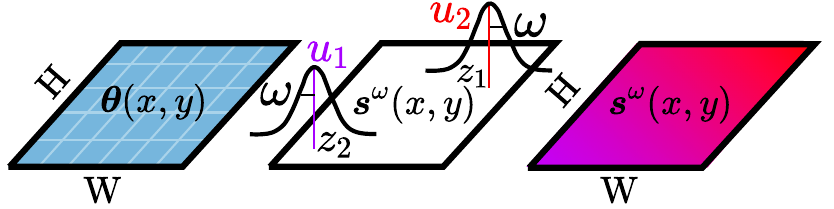}
  \end{center}
\vspace{-7.0mm}
\caption{Spatial sparsification $\vs^{\omega}$ illustrated.}
\end{wrapfigure}

\paragraph{Spatial sparsification}
To further improve the parameter efficiency, we consider parameterising spatial dimensions in a lower dimensional basis $\vsconva_{\vtheta}(c', c, \bar{x}, \bar{y}) = \sum^M_{j=1} a_j \phi^{c', c}_j(\bar{x}, \bar{y})$. Our approach is partly inspired by B-spline Lie group filters of \citep{bekkers2019b}. We follow \cite{van2023sparse} proposing standard exponential basis functions $\phi_j^{c', c}(\bar{x}, \bar{y}) = \anchor_j\exp(-\omega^2(\begin{bmatrix}\bar{x} & \bar{y}\end{bmatrix}^T - z_j)^2)$ for convolutional filters, with anchor point locations $z_j \in \R^2$ and anchor point values $\anchor_j \in \R$. The work demonstrated that lengthscales $\omega$ can allow differentiable control over equivariances (used in \cref{sec:defining-prior}) as well as practical sparsification with negligible performance loss, even if filters are already small (e.g. $3\times3$).
\vspace{-0.4em}
\begin{align}
\sum_c \sum_{x, y} \mathbf{x}(c, x, y) \vtconva(c', c, x' - x, y' - y)
&= \underbrace{\sum_c \sum_{x, y} \mathbf{x}(c, x, y) \vsconva(c', c, x' - x, y' - y)}_{\text{sparse convolution (S-CONV)}}
\vspace{-0.5em}
\label{eq:s-conv}
\end{align}
This reduces the number of parameters from $C'CS^2$ to $C'CP$ with $P$ anchor points, or $C'CP+2P$ if anchor point locations $\{ z_j \}_{j=1}^M$ (shared over channels) are also considered learnable parameters.
As weights in our proposed factored F-FC have independently factored spatial input and output domains, we can perform a similar sparsification. We write $\vsfca$ and $\vsfcb$ for spatially sparsified $\vtffca$ and $\vtffcb$:
\vspace{-0.4em}
\begin{align}
\sum_c \vtffca(c', c, x', y') \sum_{x, y}\mathbf{x}(c, x, y) \vtffcb(c', c, x, y)
&= \underbrace{\sum_c \vsfca(c', c, x', y') \sum_{x, y}\mathbf{x}(c, x, y) \vsfcb(c', c, x, y)}_{\text{sparse fully-connected (S-FC)}}
\label{eq:s-fc}
\raisetag{1.8\baselineskip}
\end{align}
reducing the parameter cost from $2C'CXY$ to $C'CP$ with $P$ total number of anchor points. Compared to convolutional filters, which already have small spatial supports, sparsification of fully-connected layers will typically result in a much larger absolute reduction of parameters, for fixed $P$. As discussed in \cref{sec:other-groups}, sparsified layers of \cref{eq:s-conv,eq:s-fc} can be extended to other groups, but require well-defined basis functions $\phi$ on groups, as explored in \citep{azangulov2022stationary}.

\subsection{Specifying the amount of equivariance in the prior}
\label{sec:defining-prior}

To allow for learnable equivariance, we explicitly specify symmetry constraints in the prior and then empirically learn them through approximate Bayesian model selection. We choose a Gaussian form $\mathcal{N}(\vzero, \sigma^2_l \mI)$ and treat variances as hyperparameters $\veta = [ \sigma_1, \sigma_2, \ldots ]$. We carefully parameterise the network, such that the prior in each layer $l$ enforces an equivariance symmetry in the following limit:
\begin{align}
\label{eq:eqprior}
\begin{split}
\sigma^2_l = 0 \implies \text{strict equivariance} \text{ (\cref{eq:equivariance-condition})}
\end{split} \hspace{1.5em} \text{and}
\begin{split}
\sigma^2_l > 0 \implies \text{relaxed equivariance} \hspace{1em}
\end{split}
\end{align}
We can think of different possible ways to parameterise neural network layers such that the conditions of \cref{eq:eqprior} are met. For instance, \cite{finzi2021residual} propose placing priors directly on the weights of both residual pathways FC+CONV $\mathcal{N}(\vtfca | \vzero, \sigma^2_l \mI) \text{ and } \mathcal{N}(\vtconva | \vzero, \bar{\sigma}^2_l \mI)$ with scalar variances $\sigma^2_l,\bar{\sigma}^2_l \in \R$ and identity matrix $\mI \in \R^{|\vtheta|\times|\vtheta|}$. Intuitively, all weights in non-equivariant paths become zero in the zero variance limit $\sigma_l{=}0$, resulting in strict equivariance. We consider the same prior for the factored parameterisation F-FC+CONV. Likewise, for sparsified layers, we can place the prior directly on anchor points $\mathcal{N}(\vu | \vzero, \sigma^2_l \mI) \text{ and } \mathcal{N}(\bar{\vu} | \vzero, \bar{\sigma}^2_l \mI)$ which will control equivariance through a similar mechanism. For S-CONV, this prior corresponds to the correlated prior of \cite{fortuin2021bayesian} if anchor points are placed exactly on the filter grid, whereas our anchor points are sparse $M{<}S^2$ and have learnable positions $z_j$. Alternatively, for sparsified S- layers, we can consider placing the prior directly on the lengthscales $\mathcal{N}(\omega | \vzero, \sigma^2_l \mI) \text{ and } \mathcal{N}(\bar{\omega}_l | \vzero, \bar{\sigma}^2_l \mI)$, as proposed in non-stationary filters \citep{van2022relaxing,van2023sparse}. Intuitively, the limit of $\sigma_l{=}0$ forces $\omega_l$ to be zero, causing non-stationary components to be constant and weights to act stationary, resulting in strict equivariance. Treating lengthscales $\vomega$ as learnable hyperparameter has the additional advantage that it allows learning of convolutional filter frequencies from data through model selection of hyperparameter $\bar{\sigma}^2_l$.

Different constructions of the prior may induce different prior probabilities in function space. Nevertheless, it can be shown that \cref{eq:eqprior} holds for all choices of prior described above. Thus, we have explicit control over symmetry with strict equivariance at the $\sigma_l {=} 0$ limiting case and relaxed layer-wise equivariance at higher variances $\sigma_l {>} 0$. Unlike prior works \citep{finzi2021residual,van2022relaxing} that require setting or tuning the prior variances $\sigma^2_l$ controlling equivariance constraints, we follow Bayesian model selection to learn them automatically from training data. 


\section{Bayesian model selection of symmetry constraints}
\label{sec:bayes-equiv}

Inspired by Bayesian methodology, we propose to empirically learn the amount of layer-wise equivariance in a neural network from training data through approximate Bayesian model selection. We define equivariances in our prior and treat the amount of symmetry as hyperparameters optimised using the marginal likelihood -- a well-understood statistical procedure known as `empirical Bayes'. Mathematically, we find a point estimate of the hyperparameters $\boldsymbol{\eta}_* {=} \argmax_{\boldsymbol{\eta}} p(\mathcal{D} | \boldsymbol{\eta})$, whilst integrating out the $P$ model parameters $\vtheta 
\in \R^P$:
\begin{align}
p(\mathcal{D} | \boldsymbol{\eta}) = \int_\vtheta p(\mathcal{D} | \vtheta, \boldsymbol{\eta}) p(\vtheta) \mathrm d \vtheta
\end{align}
The \textit{marginal likelihood} $p(\mathcal{D} | \veta)$ forms the normalising constant of the unnormalised posterior over parameters:
\begin{align}
p(\vtheta | \mathcal{D}, \veta)
= \frac{p(\mathcal{D} | \vtheta, \veta)p(\vtheta)}{p(\mathcal{D}| \veta)}
= \frac{p(\mathcal{D} | \vtheta, \veta)p(\vtheta)}{\int_{\vtheta} p(\mathcal{D}|\vtheta, \veta) p(\vtheta) \mathrm d \vtheta}
\end{align}
As the marginal likelihood is constant in $\vtheta$, it can be ignored when training neural network parameters with a regular MAP point-estimate $\vtheta_* = \argmin_{\vtheta} \mathcal{L}_{\vtheta}$:
\begin{align}
\mathcal{L}_{\vtheta} = -\log p(\mathcal{D} | \vtheta, \veta) - \log p(\vtheta)
\end{align}
To optimise hyperparameters, the marginal likelihood $p(\mathcal{D} | \veta)$ can not be ignored and becomes important again as the objective function for hyperparameters. The marginal likelihood is the integral over the very high-dimensional space of possible parameter configurations, which is intractable for large neural networks and therefore requires additional approximations in practice.

To overcome the intractability of the marginal likelihood, the Laplace approximation \citep{mackay2003information} can be used to estimate the normalising constant by first approximating the posterior by a Gaussian centered at the mode $\boldsymbol{\mu} = \vtheta_*$ and the covariance set to the local curvature $\boldsymbol{\Sigma} = \mathbf{H}_{\vtheta_*}^{-1} := -\nabla^2_{\vtheta} \mathcal{L}_{\vtheta} |_{\vtheta{=}\vtheta_*}$ at that point:
\begin{align}
p(\vtheta | \mathcal{D}, \veta) \approx q(\theta | \mathcal{D}, \veta) = \mathcal{N}(\vmu, \mSigma)
\end{align}
Integrals of Gaussians are known in closed-form, and we can thus use the normalising constant of $q$ to estimate the log marginal likelihood objective used to optimise hyperparameters $\veta_* = \argmin_{\veta} \mathcal{L}_{\veta}$:
\begin{equation}
\mathcal{L}_{\veta} = - \log \left(
\sqrt{\frac{(2 \pi)^P}{| \mH_{\vtheta_*}|}} p(\mathcal{D}, \vtheta_* | \veta) \right)
= \underbrace{- \log p(\mathcal{D} | \vtheta_*, \veta)}_{\text{NLL / Data fit}} - \underbrace{\log p(\vtheta_*) - \tfrac{P}{2} \log (2\pi) + \tfrac{1}{2} \log |\mH_{\vtheta_*}|}_{\text{Occam's factor}}
\label{eq:hyperloss}
\end{equation}
Although the Laplace approximation greatly simplifies the marginal likelihood, the Hessian determinant $|\mH_{\vtheta_*}|$ remains difficult to compute for large neural networks due to quadratic scaling in parameter count $\mathcal{O}(|\vtheta|^2)$. In practice, we therefore use a local linearisation~\citep{bottou2018optimization} resulting in the generalized Gauss-Newton (GGN) approximation, which we further approximate using a block-diagonal Kronecker-factored (KFAC) structure \cite{martens2015optimizing} (see \cref{sec:kfac}). 

The objective of \cref{eq:hyperloss} penalises model complexity enforcing a form of Occam's razor~\citep{rasmussen2000occam}. Complexity control is important when learning symmetry constraints, as previous works \citep{van2021learning,immer2021scalable} have shown that just using the regular training loss $\mathcal{L}_{\vtheta}$ w.r.t. data augmentation parameters prefers collapse into a solution without symmetry constraints resulting in fully-connected structure. The marginal likelihood overcomes this issue and can learn symmetries from data, as shown in \cite{van2018learning}. The Laplace approximated marginal likelihood $\mathcal{L}_{\veta}$ can be viewed as a form of curvature-aware minimisation and has recently been used to learn invariances in large ResNets \citep{immer2022invariance}.

\section{Kronecker-factored curvature of relaxed equivariance layers}
\label{sec:kfac}

In \cref{eq:hyperloss}, we propose the Laplace approximated log marginal likelihood $\mathcal{L}_{\veta}$ as objective to learn layer-wise equivariances from data. As Hessians scale quadratically in the number of parameters $\mathcal{O}(|\vtheta|^2)$, the loss becomes intractable for large networks. We use KFAC \citep{martens2015optimizing} to approximate the Hessian of the log likelihood $\mH_{\vtheta_\ast}$ and have the Gauss-Newton approximation $\mH$:
\begin{equation}
    {\textstyle 
    \mH_{\vtheta_\ast} \approx \mH = \sum_{n} \mH_n = \sum_{n} \mJ(\vx_n)^\top \mLambda(\vx_n) \mJ(\vx_n),
}
\end{equation}
where the Jacobians of network outputs $\vf$ with respect to parameters $\vtheta$ for data points $\vx_n$ are $[\mJ(\vx_n)]_{k,p} {=} \frac{\partial \evf_k}{\partial \evtheta_p}(\vx_n)$, and $[\mLambda(\vx_n)]_{k,g} {=} \frac{\partial \log p(y_n | \vx_n; \vf(\vx_n;\vtheta))}{\partial \evf_{k} \partial \evf_{g}}$ is the log likelihood Hessian, all evaluated at $\vtheta_\ast$. KFAC is a block-diagonal approximation of the GGN, with individual blocks $\mH_l {=} \sum_{n} \mH_{n,l} {=} \sum_{n} \mJ_l(\vx_n)^\top \mLambda(\vx_n) \mJ_l(\vx_n)$ for each layer $l$ with parameters $\vtheta_l$ and Jacobian $\mJ_l(\vx)$.

For a fully-connected layer, $\vtheta \vx$ with $\vtheta {\in} \R^{G_l \times D_l}$ and $\vx {\in} \R^{D_l}$, we can write Jacobians of a single data point w.r.t.~parameters as $\mJ_l(\vx_n)^\top {=} \va_{l,n} \otimes \vg_{l,n}$, where $\va_{l,n} {\in} \R^{D_l \times 1}$ is the input to the layer and $\vg_{l,n} {\in} \R^{G_l \times K}$ is the transposed Jacobian of $\vf$ w.r.t.~the $l$th layer output. We thus have:
\begin{align}
    \mH_l &= {\textstyle \sum_{n} \mJ_l(\vx_n)^\top \mLambda(\vx_n) \mJ_l(\vx_n)
    = \sum_{n} [\va_{l,n} \otimes \vg_{l,n}] \mLambda(\vx_n) [\va_{l, n} \otimes \vg_{l,n}]^\top} \\
    &= {\textstyle \sum_{n} [\va_{l,n} \va_{l,n}^\top] \otimes [\vg_{l, n} \mLambda(\vx_n) \vg_{l, n}^\top] 
    \approx \frac{1}{N} \Big[\sum_n \va_{l,n} \va_{l,n}^\top\Big] \otimes \big[\sum_n \vg_{l, n} \mLambda(\vx_n) \vg_{l, n}^\top\big]}.
\end{align}
KFAC only requires the computation of the two last Kronecker factors, which are relatively small compared to the full GGN block. KFAC has been extended to convolutional CONV layers in \cite{grosse2016kronecker}. We derive KFAC also for factored layers F-FC of \cref{eq:f-fc} and sparsified S-FC and S-CONV layers of \cref{eq:s-fc,eq:s-conv}, allowing use in scalable Laplace approximations to learn layer-wise equivariances from data. The essential step to extend KFAC is to write the Jacobian of the $l$th layer, $\mJ_l(\vx)$, as Kronecker product allowing data points to be factored and approximated with the last KFAC step. Our derivations of KFAC for new layers are in detail provided in \cref{app:kfac_factored,app:kfac-sparsified}.

\subsection{KFAC for factored layers}
\label{sec:kfac-factored}
To obtain the KFAC of F-FC in \cref{eq:f-fc}, we write the application of both weight parameters $\vtheta_1$ and $\vtheta_2$ as a matrix mutiplication akin to a linear layer, which allows us to apply a similar approximation as in standard KFAC. 
In particular, we first split up the operations in \cref{eq:f-fc} into first applying $\vtheta_2$ and then $\vtheta_1$.
Each of these operations can be written as a matrix product $\vtheta\vx$ with corresponding appropriate sizes of $\vtheta$ and $\vx$.
In comparison to the standard KFAC case, we have a Jacobian that can be written as
\begin{equation}
    \mJ_{\vtheta}^\top = {\textstyle \sum_c} \vx_c \otimes [\nabla_{\rvy_c} \vf]^\top = {\textstyle \sum_c} \va_c \otimes \vg_c,
\end{equation}
which means an additional sum needs to be handled.
In particular, plugging the Jacobian structure into the standard KFAC approximation, we have
\begin{align}
    \mH_{\vtheta} &\approx \textstyle{\sum_n} \mJ_\vtheta(\vx_n)^\top \mLambda(\vx_n) \mJ_\vtheta(\vx_n)
    = \textstyle{\sum_n [\sum_c \va_{n,c} \otimes \vg_{n,c}] \mLambda(\vx_n) [\sum_c \va_{n,c} \otimes \vg_{n,c}]^\top} \\
    &= {\textstyle \sum_{n,c,c'} [\va_{n,c} \va_{n,c}^\top] \otimes [\vg_{n,c'} \mLambda(\vx_n) \vg_{n,c'}^\top]}
    \approx \tfrac{1}{NC} {\textstyle [\sum_{n,c} \va_{n,c} \va_{n,c}^\top] \otimes [\sum_{n,c}\vg_{n,c} \mLambda(\vx_n) \vg_{n,c}^\top]},
\end{align}
where the first approximation is from the Hessian to the GGN and the second approximation is due to KFACs exchange of sums and products in favour of numerical efficiency.
The efficiency comes due to the fact that we have two Kronecker factors instead of a single dense matrix.
This allows to compute and optimize the marginal likelihood approximation efficiently and therefore tune $\sigma_l$ values.
In \cref{app:kfac_factored}, we derive the KFAC for $\vtheta_{1}$ and $\vtheta_2$ in detail.

\subsection{KFAC for sparsified layers}
\label{sec:kfac-sparsified}
To obtain KFAC for sparsified S-FC and S-CONV layers, we extend derivations of KFAC for respective factored F-FC layers of \cref{sec:kfac-factored} and CONV layers in \citep{grosse2016kronecker}. To compute Jacobians with respect to our new anchor points $\va$ parameters, instead of the induced weights $\vsfc$, we apply the chain rule:
\begin{align}
\label{eq:s-kfac}
\begin{split}
\frac{\partial \vf}{\partial a_j}
=
\frac{\partial \vf}{\partial \vsfc}
\frac{\partial \vsfc}{\partial a_j}
\end{split},
\text{ using the partial derivative }
\begin{split}
\frac{\partial \vsfc(c', c, x, y)}{\partial a_j} = \phi^{c', c}_j(x, y)
\end{split}
\end{align}
Consequently, we can compute KFAC for sparsified layers by projecting the appropriate dimensions of Kronecker factors of existing KFAC derivations with basis functions $\phi$ stored in the forward pass. In \cref{app:kfac-sparsified}, we derive KFAC for sparsified S-FC and S-CONV layers in detail.

\newpage
\section{Experiments}

\subsection{Toy problem: adapting symmetry to task}

The symmetries in a neural network should be task dependent. Imagine a task where the full symmetry would be too restrictive. \textit{Can our method automatically adapt to what is required by the task?} Although this ability arguably becomes most important on more complex datasets where symmetries are not strict or completely unknown, the absence of ground-truth symmetry in such scenarios makes evaluation more difficult. We, therefore, turn to a simple toy problem of which underlying symmetries are known and consider more interesting larger-scale datasets in the following sections. We slightly altered the MNIST dataset \citep{lecun1989handwritten} so that digits are randomly placed in one of four image quadrants (see example left of \cref{tab:toy-problem}) and consider the task of classifying the correct digit out of 10 classes, and the task of classifying both the digit and the quadrant it is placed in, totalling 40 classes. The first task is translation invariant, as moving a digit around does not change its class. The second task, on the other hand, can not be solved under strict translation invariance due to the dependence on location. As expected, we find in \cref{tab:toy-problem} that the CONV model outperforms the FC model on the first strictly symmetric task. Conversely, the CONV model fails to solve the second task due to the symmetry misspecification and is consequently outperformed by the FC model. The proposed method with adjustable symmetry achieves high test accuracy on both tasks.

\begin{table}[ht]
  \begin{minipage}[b]{0.15\linewidth}
  \centering
  \small{Example:}
  \par\vspace{2pt}
  \includegraphics[width=1cm]{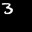}
  \par\vspace{0pt}
  \small{\textit{(3, upper left)}}
  \end{minipage}%
  \begin{minipage}[b]{0.85\linewidth}
    \centering%
    \resizebox{\linewidth}{!}{
    \begin{tabular}{l c r c c | c c c}
    \toprule
    & & & \multicolumn{2}{c}{MAP} & \multicolumn{3}{c}{Diff. Laplace} \\
    & & & FC & CONV & FC & CONV & Learned (\textbf{ours}) \\
    Symmetry & Prediction task & & 112.7 M & 0.4 M & 112.7 M & 0.4 M & 1.8 M \\
    \midrule
    Strict symmetry & \textit{(digit)} & & 95.73 & 99.38 & 97.04 & 99.23 & 98.86 \\
    Partial symmetry & \textit{(digit, quadrant)} & & 95.10 & 24.68 & 95.46 & 24.50 & 99.00 \\
    \bottomrule
    \end{tabular}
    }
    \par\vspace{0pt}
  \end{minipage}
  \caption{Preventing symmetry misspecification. Test accuracy for non-symmetric FC, strictly symmetric CONV, and learnable symmetry F-FC+CONV models on a translation invariant task and a task that can not be solved under strict translation symmetry. Unlike the FC and CONV baselines, the proposed model with learnable symmetry constraints achieves high test performance on both tasks.}
  \label{tab:toy-problem}
\end{table}

\vspace{-1em}

\subsection{Learning to use layer-wise equivariant convolutions on CIFAR-10}
\label{sec:cifar10-experiment}
\vspace{-0.5em}

\begin{wrapfigure}{R}{0.2\textwidth}
\vspace{-4.0mm}
  \begin{center}
    \includegraphics[width=1.0\textwidth]{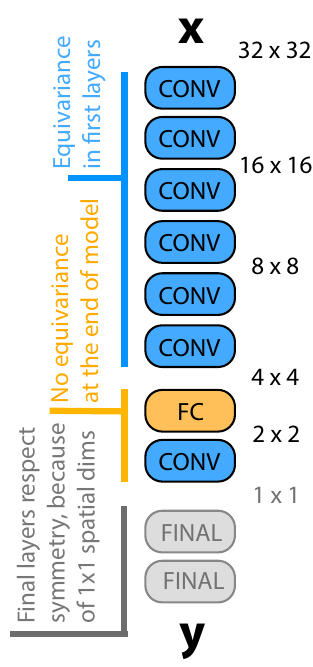}
  \end{center}
\vspace{-5.0mm}
\caption{Learned CONV+FC model.}
\label{fig:learned-layer-types}
\end{wrapfigure}

\begin{table}[b]
\vspace{-0.5em}
    \centering
    \resizebox{\linewidth}{!}{
    \begin{tabular}{l c | c | r | c c | c c c |}
    \toprule
    & Learnable & Prior & & \multicolumn{2}{c}{MAP} & \multicolumn{3}{c}{Diff. Laplace (\textbf{Ours})} \\
    Layer & equivariance & $\mathcal{N}(\cdot | \veta)$ & \# Params & Test NLL ($\downarrow$) & Test Acc. ($\uparrow$) & Test NLL ($\downarrow$) & Test Acc. ($\uparrow$) & $\mathcal{L}_{\boldsymbol{\eta}}$($\downarrow$) \\
    \midrule
    FC & & $\vtheta$ & 133.6 M & 2.423 & 64.42 & 1.319 & 52.36 & 1.896 \\
    CONV & & $\vtheta$ & 0.3 M & \textbf{1.184} & \textbf{82.81} & \textbf{0.464} & \textbf{84.07} & \textbf{1.022} \\
    FC+CONV & \cmark & $\vtheta$ & 133.9 M & 1.713 & 76.93 & \textbf{0.489} & \textbf{83.32} & \textbf{1.019} \\
    \midrule
    F-FC & & $\vtheta$ & 1.5 M & 4.291 & 50.47 & 1.307 & 53.38 & 2.119 \\
    F-FC+CONV (\textbf{Ours}) & \cmark & $\vtheta$ & 1.8 M & 1.343 & 81.61 & 
    \textbf{0.468} & \textbf{83.92} & \textbf{1.277} \\
    \midrule
    S-FC & & $\vu$  & 0.8 M & 5.614 & 50.21 & 1.496 & 46.18 & 1.926 \\
    S-CONV & & $\vu$ & 0.2 M & \textbf{0.604} & \textbf{82.53} & \textbf{0.562} & \textbf{81.29} & \textbf{1.037} \\
    S-FC+S-CONV (\textbf{Ours}) & \cmark & $\vu$  & 1.0 M & 5.485 & 58.80 & \textbf{0.538} & \textbf{81.55} & \textbf{0.978} \\
    \midrule
    S-FC & & $\vomega$  & 0.8 M & 7.783 & 49.69 & 1.319 & 52.22 & 1.813 \\
    S-CONV & & $\vomega$ & 0.2 M & \textbf{0.913} & \textbf{81.84} & \textbf{0.549} & \textbf{81.20} & \textbf{1.055} \\
    S-FC+S-CONV (\textbf{Ours}) & \cmark & $\vomega$  & 1.0 M & 2.040 & 77.97  & 0.564 & \textbf{80.63} & 1.067 \\
    \bottomrule
    \end{tabular}
    \caption{Equivariant convolutional layers CONV obtain the highest performance on image classification task when trained with regular MAP. When maximising Laplace approximated marginal likelihood, models with learnable symmetry learn from training data to `become convolutional' (see \cref{sec:cifar10-experiment} for a discussion) and obtain equivalent (within ${<}1\%$) performance as CONV models.}
    \label{tab:cifar10-results}
    }
\end{table}

Convolutional layers provide useful inductive bias for image classification tasks, such as CIFAR-10 \citep{krizhevsky2009learning}. We investigate whether our method can select this favourable convolutional structure when presented with this task. We follow \cite{finzi2021residual} and use the architecture of \cite{neyshabur2020towards} (see \cref{app:implementation}). In \cref{tab:cifar10-results}, we compare test performance when using fully-connected FC, convolutional CONV, residual pathways FC+CONV, and the proposed more parameter-efficient factorisations F-FC and F-FC+CONV and sparsified S- layers. The prior is placed on the weights $\vtheta$, anchor points $\vanchor$, or lengthscales $\vomega$, as discussed in \cref{sec:defining-prior}. As expected, CONV layers perform best under regular MAP training, followed by less constrained FC and FC+CONV layers. If we, instead, use the Laplace approximated marginal likelihood estimates, we find that learnable symmetry obtains equivalent (within <1\%) performance as the convolutional CONV models. 

\paragraph{Layers learn to become convolutional, except at the end of the network.} Upon further inspection of learned symmetries, we observe very high prior precisions $\sigma^{-2}_l {>} 10^6$ for non-equivariant paths in most layers, thus negligibly contributing to final outputs (see \cref{sec:inspecting-prior-variances}). We, therefore, conclude that layers have \textit{learned to become convolutional} from training data, which also explains why training learned equivariance with Laplace performs on par with CONV models in \cref{tab:cifar10-results}. We reach the same conclusions upon inspection of effective dimensions \citep{mackay1992bayesian} in \cref{sec:effective-number-of-parameters} and visualise learned symmetries based on the highest relative effective number of parameters in \cref{fig:learned-layer-types}. If layers did not become convolutional, this always occurred at the end of the network. This is particularly interesting as it coincides with common practice in literature. Spatial dimensions in the used architecture are halved through pooling operations, reaching 1 with sufficient or global pooling, after which features become strictly invariant and there is no difference anymore between 1x1 CONV and FC layers. Convolutional architectures in literature often flatten features and apply FC layers before this point \citep{lecun1998gradient}, breaking strict symmetry and allowing some dependence in the output on absolute position. Our findings indicate that the method can discover such favourable network designs automatically from data, outperforming regular MAP training. Furthermore, it shows that not always the same layer is selected, but different layer types are learned on a per-layer basis. 

 
\subsection{Selecting symmetry from multiple groups}

\cref{sec:other-groups} generalises the proposed method to other symmetry groups, and \cref{sec:multiple-groups} describes layer-wise equivariance selection from a set of multiple symmetry groups or layer types. In this experiment, we consider adding discrete 90-degree rotations to the architecture F-FC+CONV+GCONV. In \cref{tab:multiple-groups}, we compare the test performance with MAP and approximated marginal likelihood training on versions of MNIST and CIFAR-10 datasets (details in \cref{app:datasets}). We compensate for the added parameters resulting from the rotationally equivariant GCONV path by reducing channel sizes of individual paths by a factor of 5 ($\alpha{=}10$ to $\alpha{=}2$, see \cref{app:implementation}). Still, we are able to obtain 80\% test accuracy on CIFAR-10 when trained with approximated marginal likelihood. Upon analysis of learned prior variances (\cref{sec:inspecting-prior-variances}) and effective number of parameters (\cref{sec:effective-number-of-parameters}), we observe a positive correlation between learned symmetries and augmentations applied to the datasets trained on.

\vspace{-0.5em}
\begin{table}[h]
    \centering
    \resizebox{\linewidth}{!}{
    \begin{tabular}{r c c | c c | c c c r r r r r r }
    \toprule
    & \multicolumn{2}{c}{} & \multicolumn{2}{c}{MAP} & \multicolumn{3}{c}{Learned with Differentiable Laplace (\textbf{ours})} &  \multicolumn{6}{c}{Rel. Effective Num. of Param. }] \\
    Dataset & & \# Params & Test NLL ($\downarrow$) & Test accuracy ($\uparrow$) & Test NLL ($\downarrow$) & Test accuracy ($\uparrow$) & Approx. MargLik ($\downarrow$) & \multicolumn{2}{c}{FC (\%)} &  \multicolumn{2}{c}{CONV (\%)} &  \multicolumn{2}{c}{GCONV (\%)} \\
    \midrule
    MNIST & & 1.2 M & 0.172 & 97.59 & 0.023 & \textbf{99.21} & 0.328 & 
10 & \textcolor{gray}{(0-46)} & 15 & \textcolor{gray}{(0-98)} & 75 & \textcolor{gray}{(1-100)} \\
    Translated MNIST & & 1.2 M & 0.812 & 90.78 & 0.053 & \textbf{98.27} & 0.216 & 
0 & \textcolor{gray}{(0-0)} & 23 & \textcolor{gray}{(0-99)} & 77 & \textcolor{gray}{(1-100)} \\
    Rotated MNIST & & 1.2 M & 0.819 & 91.02 & 0.136 & \textbf{95.55} & 0.896 &
8 & \textcolor{gray}{(0-20)} & 8 & \textcolor{gray}{(0-47)} & 83 & \textcolor{gray}{(47-100)} \\
    \midrule
    CIFAR-10 & & 1.2 M & 3.540 & 68.33 & 0.552 & \textbf{80.94} & 0.926 &
0 & \textcolor{gray}{(0-1)} & 44 & \textcolor{gray}{(0-99)} & 56 & \textcolor{gray}{(0-100)} \\
    Rotated CIFAR-10 & & 1.2 M & 5.953 & 48.30 & 1.236 & \textbf{55.68} & 1.630 &
4 & \textcolor{gray}{(0-22)} & 14 & \textcolor{gray}{(0-41)} & 82 & \textcolor{gray}{(58-99)} \\
    \bottomrule
    \end{tabular}
    }
    \caption{Selecting from multiple symmetry groups. Negative log likelihood (NNL) and Laplace learned symmetries measured by \textit{mean (min-max)} relative effective number of parameters over layers.}
    \label{tab:multiple-groups}
\end{table}
\vspace{-0.5em}

\vspace{-1em}
\section{Discussion and Conclusion}

This work proposes a method to automatically learn layer-wise equivariances in deep learning using gradients from training data. This is a challenging task as it requires both flexible parameterisations of symmetry structure and an objective that can learn symmetry constraints. We improve upon existing parameterisations of relaxed equivariance to remain practical in the number of parameters. We learn equivariances through Bayesian model selection, by specifying symmetries in the prior and learning them by optimising marginal likelihood estimates. We derive Kronecker-factored approximations of proposed layers to enable scalable Laplace approximations in deep neural networks.

The approach generalises symmetry groups and can be used to automatically determine the most relevant symmetry group from a set of multiple groups. We rely on relaxing symmetries and learning the amount of relaxation, where strict equivariance forms a limiting case. In doing so, the method does require a list of symmetry groups and associated group representations that are considered, and learns to ignore, partially use, or strictly obey symmetry constraints. Yet, we argue this is a huge step forward compared to always using a single, strict symmetry. We hope that allowing artificial neural networks to automatically adapt structure from training data helps to leverage existing geometry in data better and reduce the reliance on trial-and-error in architecture design.

\newpage
\bibliographystyle{plainnat}
\bibliography{neurips_2023}

\appendix

\newpage
\section{Generalisation to other groups}
\label{sec:other-groups}

The main text describes layers with strict or relaxed equivariance to the translation group because this is the most commonly used symmetry in deep learning architectures used in classical convolutional layers. In this section, we show that all layers can be generalised to other symmetry groups $G$, where 2-dimensional discrete translations correspond to the special case of $G = \Z^2$. In this case any two group elements $(x, y) = g \in G = \Z^2$ and inverses become subtraction $g^{-1}g' = (x' - x, y' - y)$.

On groups, we consider input and output feature maps on a group $\vy : [0, B] \times [0, C] \times G$ and $\vx : [0, B] \times [0, C] \times G$, which can be achieved through group lifting \citep{kondor2018generalization}. We generalise the linear layer of \cref{eq:linear-layer} on a group $G$ as:
\begin{align}
\mathbf{y}(c', g') &= \sum_c \sum_{g \in G} \vtfc(c', c, g', g) \mathbf{x}(c, g) 
\label{eq:linear-layer-group}
\end{align}
Generalising regular convolutional layer to a group $G$ becomes the group equivariant convolution proposed in \cite{cohen2016group}:
\begin{align}
\label{eq:conv-group}
\mathbf{y}(c', g') &= \sum_c \sum_{g} \mathbf{x}(c, g) \vtconv(c', c, g^{-1}g')
\end{align}
The residual pathway of \cref{eq:rpp-group} was originally in \cite{finzi2021residual} for general groups $G$. It can be seen as a factorisation $\vtfc = \vtfca + \vtconva$ where $\vtfca : (c', c, g', g) \mapsto \R$ and $\vtconva : (c', c, g) \mapsto \R$:
\begin{align}
\label{eq:rpp-group}
\mathbf{y}(c', g')
&= \sum_c \sum_{g} \mathbf{x}(c, g) \vtfca(c', c, g', g) + \sum_c \sum_{g} \mathbf{x}(c, g) \vtconva(c', c, g^{-1}g')
\end{align}
The factorisation of \cref{eq:f-fc} proposed to further reduce the number of parameters generalises as:
\begin{align}
\label{eq:f-fc-group}
\sum_c \sum_{g} \mathbf{x}(c, g) \vtfca(c', c, g', g)
&= \sum_c \vtffca(c', c, g') \sum_{g}\mathbf{x}(c, g) \vtffcb(c', c, g) 
\end{align}
where $\vtffca : (c', c, g') \mapsto \R$ and $\vtffcb : (c', c, g) \mapsto \R$.
The sparse fully-connected layer of \cref{eq:s-fc} generalises to groups $G$ as:
\begin{align}
\sum_c \sum_{g} \mathbf{x}(c, g) \vtconva(c', c, g^{-1}g')
&= \sum_c \sum_{g} \mathbf{x}(c, g) \vsconva(c', c, g^{-1}g')
\end{align}
with stationary filters $\vtconva : (c', c, u) \mapsto \R$ taking group elements $u{=}g^{-1}g'\in G$ as argument. The sparse convolutional layer of \cref{eq:s-conv} becomes:
\begin{align}
\sum_c \vtffca(c', c, g') \sum_{g}\mathbf{x}(c, g) \vtffcb(c', c, g)
&= \sum_c \vsfca(c', c, g') \sum_{g}\mathbf{x}(c, g) \vsfcb(c', c, g)
\end{align}
where $\vsfca : ... $ and $\vsfcb : ... $. In this case, the bases become functions on the group $\phi : G \to \R$, as have been explored in \cite{azangulov2022stationary}.

%


\section{Selecting symmetry from multiple groups}
\label{sec:multiple-groups}

Consider a set of $M$ groups $G_1, G_2, \ldots, G_M$, then we can factor multiple groups as follows: 
\begin{align}
\label{eq:multiple-groups}
\mathbf{y}(c', g')
&= \sum_{i=1}^M \sum_c \sum_{g \in G_i} \mathbf{x}(c, g) \vtfca(c', c, g', g) +
\sum_{i=1}^M \sum_c \sum_{g \in G_i} \mathbf{x}(c, g) \vtconva(c', c, g^{-1}g')
\end{align}
strictly generalising residual pathways of \cref{eq:rpp-group} to multiple groups $\{ G_i \}_{i=1}^M$. The other layers can equivalently be extended to multiple groups.

Similar to the original residual pathway paper \citep{finzi2021residual}, we could also write linear mappings as matrix $\mW$ and consider a set of groups $G_1, G_2, \ldots, G_M$ with representations and find a corresponding set of equivariant bases $\mB_1, \mB_2, \ldots \mB_M$  \citep{finzi2021practical}, in which case we could write the layer of \cref{eq:multiple-groups} equivalently as a sum of these linear maps:
\begin{align}
\text{vec}(\mW) = \sum_{i=1}^G \mathbf{B}_i \vu_i + \vv
\end{align}
The latter notation makes it possibly more clear that the layer forms a sum of linear layers that span subspaces associated with different equivariance constraints.

\section{Extensions of Kronecker-factored approximate curvature (KFAC)}
We briefly review KFAC for fully-connected neural networks and then extend it to the layers proposed in this work, i.e., factored layers and sparsified layers.
Further, we give an extension for group convolutional layers.
The goal of KFAC is to approximate the Hessian of the log likelihood, $\mH_{\vtheta_\ast}$, which is the overall loss Hessian from which the simple Hessian of the prior is subtracted.
We have for the Gauss-Newton approximation $\mH$:
\begin{equation}
    \mH_{\vtheta_\ast} \approx \mH = \sum_{n} \mH_n = \sum_{n} \mJ(\vx_n)^\top \mLambda(\vx_n) \mJ(\vx_n),
\end{equation}
where $\mJ(\vx_n)$ are the Jacobians of neural network output $\vf$ with respect to parameters $\vtheta$ for a given data point $\vx_n$, i.e., $[\mJ(\vx_n)]_{k,p} = \frac{\partial \evf_k}{\partial \evtheta_p}(\vx_n)$, and $[\mLambda(\vx_n)]_{k,g} = \frac{\partial \log p(y_n | \vx_n; \vf(\vx_n;\vtheta))}{\partial \evf_{k} \partial \evf_{g}}$, which is the Hessian of the log likelihood with respect to the functional output of the neural network and is for common likelihoods independent of the label~\citep{martens2020new}.

KFAC approximates the GGN in two efficient ways: first, the approximation is conducted in blocks, each corresponding to a single layer $l$, and, second, the GGN of each block is approximated as a Kronecker product.
We denote a block corresponding to the $l$th layer as $\mH_l$, which is the GGN of the parameter $\vtheta_l$, and has the Jacobian $\mJ_l(\vx)$.
Then, we have
\begin{equation}
    \mH_l = \sum_{n} \mH_{n,l} = \sum_{n} \mJ_l(\vx_n)^\top \mLambda(\vx_n) \mJ_l(\vx_n),
\end{equation}
For a fully-connected layer, $\vtheta \vx$ with $\vtheta \in \R^{G_l \times D_l}$ and $\vx \in \R^{D_l}$, we can write the Jacobian of a single data point w.r.t.~parameters as $\mJ_l(\vx_n)^\top = \va_{l,n} \otimes \vg_{l,n}$, where $\va_{l,n} \in \R^{D_l \times 1}$ is the input to the layer and $\vg_{l,n} \in \R^{G_l \times K}$ is the transposed Jacobian of $\vf$ w.r.t.~the output of the $l$th layer.
KFAC can then be derived using the following equalities and lastly approximation:
\begin{align}
    \mH_l &= \sum_{n} \mJ_l(\vx_n)^\top \mLambda(\vx_n) \mJ_l(\vx_n)
    = \sum_{n} [\va_{l,n} \otimes \vg_{l,n}] \mLambda(\vx_n) [\va_{l, n} \otimes \vg_{l,n}]^\top \\
    &= \sum_{n} [\va_{l,n} \va_{l,n}^\top] \otimes [\vg_{l, n} \mLambda(\vx_n) \vg_{l, n}^\top] 
    \approx \frac{1}{N} \left[\sum_n \va_{l,n} \va_{l,n}^\top\right] \otimes \left[\sum_n \vg_{l, n} \mLambda(\vx_n) \vg_{l, n}^\top\right].
\end{align}
The resulting KFAC just requires to compute the two last Kronecker factors, which are relatively small compared to the full GGN block.
To extend KFAC to the proposed layers, the essential step is to write the Jacobian of the $l$th layer, $\mJ_l(\vx)$, as a Kronecker product so it can be factored for each data point and then approximated with the last KFAC step.

\subsection{KFAC for factored layers}
\label{app:kfac_factored}
Factored layers have two parameters $\vtheta_1, \vtheta_1$, each of which will result in a KFAC approximation of its respective curvature.
First, we split up the definition from \cref{eq:f-fc}:
\begin{align}
\rvy(c', x', y') &= \sum_c \sum_{x, y} \mathbf{x}(c, x, y) \vtfca(c', c, x', y', x, y) \\
&= \sum_c \vtffca(c', c, x', y') \sum_{x, y}\mathbf{x}(c, x, y) \vtffcb(c', c, x, y) \\
&= \sum_c \vtffca(c', c, x', y') \vx_1(c', c),
\end{align}
which defines $\vx_1(c', c)$ as intermediate value, i.e., it is input to $\vtffca$ and output of operation $\vtffca$.
Both individual operations can be equivalently written similar to linear layers by modifying the inputs. 

For $\vtffcb$, we write $\vtffcb \in \R^{c' \times cxy}$ and $\vx \in \R^{cxy \times c}$ which is given by repeating $\vx$ across the $c$ dimensions along the diagonal.
We then have $\vx_1^{(o)} = \vtffcb \vx \in \R^{c' \times c}$.
For $\vtffca$, we have similarly $\vtffca \in \R^{x'y' \times cc'}$ and $\vx_1^{(i)} \in \R^{cc' \times c'}$, which is a block-diagonal constructed from the previous output $\vx_1^{(o)}$, i.e., $\vtffca$ is applied individually per $c'$.
We have $\rvy \in \R^{x'y' \times c'} = \vtffca \vx_1^{(i)}$.
To obtain KFAC for the two parameters, we need to write out the Jacobian as a Kronecker product of input data and output gradient.
The input data are given by the expanded $\vx$ and transposed output Jacobians are $d\rvy \in \R^{x'y'c' \times K}$ and $d\vx \in \R^{c'c\times K}$.
This can be done as follows:
\begin{align}
    \mJ_{\vtffcb} = \sum_{c} d\vx^\top_c \otimes \vx_c, \quad \quad \textrm{and} \quad \quad
    \mJ_{\vtffca} = \sum_{c'} d\rvy^\top_{c'} \otimes (\vx_1^{(i)})_{c'}
\end{align}
where $c$ and $c'$ denote the index of the repeated dimension, respectively.
Prototypically for $\vtheta_2$, defining $\va_{l,n,c}$ as $\vx_c$ and $\vg_{l,n,c}$ as $d\vx_c^\top$, we have the following KFAC approximation:
\begin{align}
    \mH_{l} &= \sum_n \left[\sum_c \va_{l,n,c} \vg_{l,n,c}\right]\otimes \left[\sum_c \va_{l,n,c} \mLambda(\vx_n) \vg_{l,n,c}\right] \\
    &\approx \frac{1}{NC} \sum_n \sum_c \left[\va_{l,n,c} \va_{l,n,c}^\top\right] \otimes \left[\vg_{l,n,c} \mLambda(\vx_n) \vg_{l,n,c}^\top\right],
\end{align}
which equivalently holds for $\vtheta_1$ by replacing $c$ with $c'$, i.e., the number of input channels with output channels.

\subsection{KFAC for sparsified layers}
\label{app:kfac-sparsified}

For sparsified factored layers S-FC, we extend the K-FAC approximation for factored layers. We follow the same derivation as in \cref{app:kfac_factored} to obtain Jacobians $\mJ_{\vsfca}$ and $\mJ_{\vsfcb}$ with respect to weights induced by the basis functions $\vsfca$, $\vsfcb$:
\begin{align}
    \mJ_{\vsfca} = \sum_{c} d\vx^\top_c \otimes \vx_c, \quad \quad \textrm{and} \quad \quad
    \mJ_{\vsfcb} = \sum_{c'} d\rvy^\top_{c'} \otimes (\vx_1^{(i)})_{c'}
\end{align}
In sparsified layers, $\vsfca$ and $\vsfcb$ are functions and not part of the model parameters anymore. Instead, we are looking for the Jacobians $\mJ_{\vanchor_1}$ and $\mJ_{\vanchor_2}$ with respect to the sparser set of anchor point weights $\vanchor_1$ and $\vanchor_2$, which we can find by applying the chain rule:
\begin{align}
\begin{split}
\frac{\partial \vf}{\partial \vanchor_1} = \frac{\partial \vf}{\partial \vsfca} \frac{\partial \vsfca}{\partial \vanchor_1}
\end{split}\text{       , and    }
\begin{split}
\frac{\partial \vf}{\partial \vanchor_2} = \frac{\partial \vf}{\partial \vsfcb} \frac{\partial \vsfcb}{\partial \vanchor_2}
\end{split}
\end{align}
where the partial derivatives are given by basis functions ${}^{1}\phi$ and ${}^{2}\phi$ associated to $\vanchor_1$ and $\vanchor_2$:
\begin{align}
\begin{split}
\frac{\partial \vsfca(c', c, x, y)}{\partial (\vanchor_1)_j} = {}^{1}\phi^{c', c}_j(x, y)
\end{split}\text{       , and    }
\begin{split}
\frac{\partial \vsfcb(c', c, x, y)}{\partial (\vanchor_1)_j} = {}^{2}\phi^{c', c}_j(x, y)
\end{split}
\end{align}
As basis functions form partial derivatives from $\vsfca$ and $\vsfcb$ to $\vanchor_1$ and $\vanchor_2$, respectively, we can use them to project the factors used in \cref{sec:factor-sparsify} in terms of the new sparsified parameters.

For sparsified convolutional layers S-CONV, we extend the KFAC approximation for convolutional layers derived in \citep{grosse2016kronecker}. Similarly to above, we use the existing derivation to approximate KFAC in terms of Jacobians w.r.t. induced convolutional filters $\mJ_{\vsconva}$. Filters $\vsconva$ in S-CONV layers are not part of the model parameters anymore and we need to find Jacobians $\mJ_{\vanchor}$ with respect to anchor point weights $\bar{\vanchor}$, which we can find by applying the chain rule:
\begin{align}
\begin{split}
\frac{\partial \vf}{\partial \vanchor} = \frac{\partial \vf}{\partial \vsconva} \frac{\partial \vsfcb}{\partial \vanchor_2}
\end{split}\text{, with}
\begin{split}
\frac{\partial \vsconva(c', c, u, v)}{\partial \bar{\vanchor}_j} = \bar{\phi}_j^{c', c}(u, v)
\end{split}
\end{align}
Similar to S-FC layers, we can project Kronecker factors of KFAC for CONV layer into the right sparser parameter space by multiplying the appropriate dimensions with basis function evaluations.

\newpage
\section{Inspecting learned layer-wise structure.}

\begin{table}[ht]
    \centering
    \resizebox{\linewidth}{!}{
    \begin{tabular}{c|r r|r r|r r|r r|c}
    \toprule
    & \multicolumn{2}{c}{Prior precisions} & \multicolumn{2}{c}{Weight norms} &  \multicolumn{2}{c}{Normalised Effective Num. of Param.} & \multicolumn{2}{c}{Relative Effective Num. of Param.} & \\
    & \multicolumn{2}{c}{$\uparrow$ implies 'off'} & \multicolumn{2}{c}{$\downarrow$ implies 'off'} &  \multicolumn{2}{c}{$\downarrow$ implies 'off'} & \multicolumn{2}{c}{$\downarrow$ implies 'off'} & Effective \\
    Layer $l$ & FC $\sigma_l^{-2} $ & CONV $\bar{\sigma}_l^{-2}$ & FC $||\vtfca_l||_2^2$ & CONV $||\vtconv_l||^2_2$ & FC $\frac{\gamma_l}{\gamma_l + \bar{\gamma}_l}$ & CONV $\frac{\gamma_l}{\gamma_l + \bar{\gamma}_l}$ & FC $\frac{\gamma_l/P_l}{\gamma_l/P_l + \bar{\gamma}_l/\bar{P}_l}$ & CONV $\frac{\bar{\gamma}_l/\bar{P}_l}{\gamma_l/P_l + \bar{\gamma}_l/\bar{P}_l}$ & Layer type \\
    \midrule
0 & 3588348.50000000 & 0.07513507 & 0.00000000 & 10.74257360 & 0.00000002 & 0.80731778 & 0.00000003 & 0.99999997 & CONV \\
1 & 2750749.50000000 & 1.26366782 & 0.00000000 & 0.55490241 & 0.00000007 & 0.70107951 & 0.00000010 & 0.99999990 & CONV \\
2 & 5898884.50000000 & 14.50690365 & 0.00000000 & 0.04562404 & 0.00000014 & 0.66181451 & 0.00000021 & 0.99999979 & CONV \\
3 & 22690960.00000000 & 230.75096130 & 0.00000000 & 0.00216724 & 0.00000089 & 0.50004517 & 0.00000178 & 0.99999822 & CONV \\
4 & 29633508.00000000 & 5438.10449219 & 0.00000000 & 0.00007521 & 0.00000772 & 0.40905216 & 0.00001888 & 0.99998112 & CONV \\
5 & 35631228.00000000 & 3797.48779297 & 0.00000000 & 0.00003772 & 0.00000379 & 0.14326709 & 0.00002643 & 0.99997357 & CONV \\
6 & 635.57513428 & 31321916.00000000 & 0.00001135 & 0.00000000 & 0.00721232 & 0.00000522 & 0.99927627 & 0.00072373 & FC \\
7 & 35582056.00000000 & 1.75766361 & 0.00000000 & 0.00259477 & 0.00000009 & 0.00456014 & 0.00001874 & 0.99998126 & CONV \\
    \bottomrule
    \end{tabular}
    }
    \caption{Analysis of learned layer types after training FC+CONV model on CIFAR-10. Reported prior precision, weight norms, effective dimension and relative effective dimension metrics. On this task, the model learns to use CONV layers, except for a FC layer at the end of the model.}
    \label{tab:inspecting-learned-types}
\end{table}

\subsection{Learned prior variances}
\label{sec:inspecting-prior-variances}

To learn symmetries, we follow Bayesian model selection by specifying symmetry in the prior and learning it by optimising marginal likelihood estimates. For each layer, we consider a Gaussian prior with prior precisions $\frac{1}{\sigma^2}$ controlling the amount of equivariance. Most notably, in the limit of $\lim_{c\to\infty} \frac{1}{\sigma^2} = c$ we have strict equivariance. We can inspect prior precisions of both the CONV path $\bar{\sigma}^{-2}$ and the FC path $\sigma^{-2}$, where high prior precisions can intuitively interpret as the layer being `switched off'. Prior precisions learned for the CONV+FC model are shown on the left in \cref{tab:inspecting-learned-types}. Although prior precisions may not always be directly interpretable (see \cref{sec:effective-number-of-parameters}, the learned prior precisions indicate that the model learns convolutional structure for most layers, except for layer 7 at the end, which we argue is a desirable solution as hypothesised in \cref{sec:cifar10-experiment}.

%

\subsection{Effective number of parameters}
\label{sec:effective-number-of-parameters}

\citet{mackay1992bayesian} defines the effective number of parameters $\gamma$ using the Laplace approximation to the posterior.
In particular, it measures how well a parameter is determined relative to its prior.
\citet{mackay1992bayesian} defines it for a model with Gaussian prior $\mathcal{N}(0, \alpha^{-1} \mI_P)$ over $P$, whose posterior covariance is given by $\mSigma$.
We then have
\begin{equation}
    \gamma = P - \alpha\Tr(\mSigma) = P - \alpha \Tr(\mH + \alpha \mI_P)^{-1} = P - \sum_{p=1}^P \frac{\alpha}{\lambda_p + \alpha}
    = \sum_{p=1}^P \frac{\lambda_p}{\lambda_p + \alpha},
\end{equation}
where $\mH$ is the Hessian at the posterior mode.
However, this therefore requires the full posterior covariance $\mSigma$ due to the Laplace approximation.
In our case, we further have a layer-wise KFAC approximation over the individual layers $l$ with corresponding parameters.
Our models share the following setup: we have a prior variances $\sigma_l^2$ and $\bar{\sigma}_l^2$ per layer, each corresponding to the fully connected and convolutional path, respectively.
Further, let $P_l$ and $\bar{P}_l$ be the number of distinct parameters in the respective layers.
Our posterior approximation is block-diagonal and Kronecker-factored, i.e., we have that $\mH$ is a block-diagonal constructed from $\mH_l$ and $\bar{\mH}_l$ for all layers $l$ and each can be written as $\mH_l \approx \mQ_l \otimes \mW_l$.
Therefore, we have
\begin{align}
    \gamma &= \sum_{l=1}^L P_{l} - \sigma_l^{-2} \Tr(\mH_l + \sigma_l^{-2})^{-1} + \bar{P}_{l} - \bar{\sigma}_l^{-2} \Tr(\bar{\mH}_l + \bar{\sigma}_l^{-2})^{-1} \\
    &= \sum_{l=1}^L \underbrace{\sum_{w,q} \frac{\lambda_{l,w} \lambda_{l,q}}{\lambda_{l,w} \lambda_{l,q} + \sigma_l^{-2}}}_{=\gamma_l} + \underbrace{\sum_{w,q} \frac{\bar{\lambda}_{l,w} \bar{\lambda}_{l,q}}{\bar{\lambda}_{l,w} \bar{\lambda}_{l,q} + \bar{\sigma}_l^{-2}}}_{=\bar{\gamma}_l},
\end{align}
where $\lambda_{l,w}$ denotes the eigenvalue of $\mW_l$ and similarly for $\lambda_{l,q}$.
The simplification is due to the eigenvalues of the Kronecker product being the outer product of its factors eigenvalues. Further, denote $\gamma_l$ and $\bar{\gamma}_l$ the effective number of parameters for FC and CONV layer at $l$. For FC and CONV, respectively, we define the `normalised effective number of parameters' as $\frac{\gamma_l}{P_l}$ and $\frac{\bar{\gamma}_l}{\bar{P}_l}$ and the `relative effective number of parameters' as $\frac{\gamma_l/P_l}{\gamma_l/P_l + \bar{\gamma}_l / \bar{P}_l}$ and $\frac{\bar{\gamma}_l/\bar{P}_l}{\gamma_l/P_l + \bar{\gamma}_l/\bar{P}_l}$. The definitions generalise trivially if more than these two layer types are being considered.

%

\newpage
\section{Comparison of existing approaches}

\begin{table}[ht]
    \centering
    \resizebox{\linewidth}{!}{
    \begin{tabular}{r|c c c c c c }
    \toprule
    & Symmetry & Layer-wise & Automatic & No validation & No explicit & Sparse layer-wise \\
    Method & learning & equivariance & objective & data & regulariser & symmetry \\
    \midrule
    \cite{zhou2019bayesnas}     & \checkmark & \checkmark & \checkmark & \checkmark & \checkmark & \\
    \cite{benton2020learning}     & \checkmark & & & \checkmark & & \\
    \cite{zhou2020meta}           & \checkmark & \checkmark & \checkmark & & \checkmark & \\
    \cite{romero2021learning}     & \checkmark & \checkmark & & & \checkmark & \\
    \cite{finzi2021residual}      & \checkmark & \checkmark & & \checkmark & & \\
    \cite{dehmamy2021automatic}   & \checkmark & \checkmark & & \checkmark & \checkmark & \checkmark \\
    \cite{immer2022invariance}    & \checkmark & & \checkmark & \checkmark & \checkmark & \\
    \cite{yeh2022equivariance} & \checkmark & \checkmark & \checkmark & & \checkmark & \\
    \cite{maile2022architectural} & \checkmark & \checkmark & \checkmark & & \checkmark & \\
    \cite{van2022relaxing}        & \checkmark & \checkmark & & \checkmark & & \\
    \cite{yang2023generative} & \checkmark & \checkmark & & \checkmark & & \\
    \cite{van2023sparse}          & \checkmark & \checkmark & & \checkmark & & \checkmark \\
    (\textbf{Ours})   & \pmb{\checkmark} & \pmb{\checkmark} & \pmb{\checkmark} & \pmb{\checkmark} & \pmb{\checkmark} & \pmb{\checkmark} \\
    \bottomrule
    \end{tabular}
    }
    \caption{Overview of existing methods for symmetry discovery.}
    \label{tab:learnedexisting-methods}
\end{table}

\paragraph{Symmetry learning: } In this comparison, we include methods that learn symmetries in the context of deep neural networks, including both learnable invariances and equivariances.

\paragraph{Layer-wise equivariance:} Invariances can often be parameterised by averaging data augmentations after forward passes through a model. Layer-wise equivariances can be more difficult to parameterise, as this places constraints on intermediary features and, therefore, relies on adaptions inside the actual architecture of the model.

\paragraph{Automatic objective:} Symmetries enforce constraints on the functions a network can represent, which makes it hard to learn them with regular maximum likelihood training objectives that rely on data fit. To avoid collapse into solutions with the least (symmetry) constraints, some methods rely on explicit regularisation to enforce symmetry. Although this approach has been successful in some cases \citep{benton2020learning,finzi2021residual,van2022relaxing}, issues with the approach have also been noted and discussed in \cite{immer2022invariance}. The most important critique is that direct regularisation of symmetry often depends on the chosen parameterisation and introduces additional hyperparameters that need tuning. Automatic objectives learn symmetries from training data without introducing additional hyperparameters that need tuning.

\paragraph{No validation data:} No validation data is used to select or learn symmetries. Most notably, these are methods that use differentiable validation data as an objective to learn hyperparameters. We also include methods in this category that initialise networks at strict symmetry and use validation data as an early-stopping criterion, in case no other encouragement is added in the training objective - assuming that early-stopping, in this case, becomes the mechanism that prevents collapse into non-symmetric solutions.

\paragraph{Sparse layer-wise symmetry:} We deem methods that fall within a 10-fold increase of number of parameters to parameterise relaxed symmetry constraints. Although this might seem like a lot, this distinguishes methods that relax symmetry by considering fully-flexible linear maps, which in practice often leads to an increase of more than 100 times.

\newpage
\section{Implementation details}
\label{app:implementation}

\subsection{Network architecture}

\cref{tab:architecture} describes the used architecture for all experiments. The design is adapted from the convolutional architecture used in \citep{neyshabur2020towards}. We use strict equivariant subsampling described in \cref{sec:subsampling} as POOL and use element-wise $\text{ReLU}(x){=}\max(0, x)$ activation functions. With convolutional layers CONV, the architecture is strictly equivariant. In some experiments the CONV layers are replaced to relax strict layer-wise symmetry constraints.

\begin{table}[H]
    \centering
    \resizebox{0.7\linewidth}{!}{
    \begin{tabular}{l | l | l}
    \toprule
    layer & input & output \\
    \midrule
CONV($3\times3$) + ReLU & $B \times C_{\text{in}} \times 32 \times 32$ & $B \times \alpha \times 32 \times 32$ \\
CONV($3\times3$) + POOL + ReLU & $B \times \alpha \times 32 \times 32$ & $B \times 2\alpha \times 16 \times 16$ \\
CONV($3\times3$) + ReLU & $B \times \alpha \times 16 \times 16$ & $B \times 2\alpha \times 16 \times 16$ \\
CONV($3\times3$) + POOL + ReLU & $B \times 2\alpha \times 16 \times 16$ & $B \times 4\alpha \times 8 \times 8$ \\
CONV($3\times3$) + ReLU & $B \times 4\alpha \times 16 \times 16$ & $B \times 4\alpha \times 8 \times 8$ \\
CONV($3\times3$) + POOL + ReLU & $B \times 4\alpha \times 8 \times 8$ & $B \times 8\alpha \times 4 \times 4$ \\
CONV($3\times3$) + POOL + ReLU & $B \times 8\alpha \times 4 \times 4$ & $B \times 8\alpha \times 2 \times 2$ \\
CONV($3\times3$) + ReLU & $B \times 16\alpha \times 2 \times 2$ & $B \times 16\alpha \times 1 \times 1$ \\
CONV($1\times1$) + ReLU & $B \times 16\alpha \times 1 \times 1$ & $B \times 64\alpha \times 1 \times 1$ \\
CONV($1\times1$) & $B \times 64\alpha \times 1 \times 1$ & $B \times C_{\text{out}} \times 1 \times 1$ \\
    \bottomrule
    \end{tabular}
}
    \caption{Used architecture with $\alpha{=}10$.}
    \label{tab:architecture}
\end{table}

\subsection{Group subsampling and maintaining strict equivairance.}
\label{sec:subsampling}
The core idea of this work is to utilise parameterisations that have strict equivariance symmetries as controllable limiting case and learning the relative importance of the symmetry constraint through Bayesian model selection. In this framework, we encode the limiting case of strict equivariance in the prior through use of (group) convolutions with circular padding and pointwise non-linearities, which respect the equivariance property. It is known, however, that subsampling operations such as maxpooling and strided convolutions present in most convolutional architectures, including our baseline \citep{neyshabur2020towards}, do not respect equivariance strictly. To overcome this issue, we might consider coset-pooling \citep{cohen2016group}, but this looses locality of the feature maps and is therefore typically only used at the end of the architecture. Alternatives, such as \cite{xu2021group}, introduce an additional subsampling dimension which requires memory and complicates implementation. The implementation of our method is already involved because Kronecker-factored curvature approximations are to date not naively supported in most deep learning frameworks. We therefore follow the more pragmatic approach of \cite{chaman2021truly} and subsample feature maps by breaking them up into polyphase components selecting the component with the highest norm (we use $l_{\infty}$-norm). This is easy to implement and maintains both strict equivariance and locality of feature maps. Furthermore, the baseline is not negatively affected by this change as regular MAP performance of the CONV model improved by 1-2 \% percentage points in test accuracy.

\subsection{Locality of support, separable group convolutions, pointwise group convolutions.}

In deep learning literature, convolutional filters commonly only have a very small local support (e.g. $5\times5$, $3 \times 3$ or even $1\times1$). This is efficient and induces a locality bias that has empirically been found beneficial for many tasks. Interestingly, group equivariant convolutional filters are rarely locally supported and often defined on the entire group. Recent work by \cite{knigge2022exploiting}, explores factoring filters by generalising the concept of (spatially) separable convolutions to (sub)groups. Spatial separability implies the filter factorisation $\vtconv(c', c, x, y) = \vtconv(c', c, x)\vtconv(c', c, y)$. Applying the same concept to separate rotation and translation subgroups for roto-translation equivariance with $(x, y, \theta) \in G = \Z^2 \rtimes p4$ factors filters over subgroups $\vtconv(c', c, x, y, \theta) = \vtconv(c', c, x, y)\vtconv(c', c, \theta)$. We take this one step further and propose filters that are only pointwise for specific subgroups, which can be seen as a special case of separable group convolution with $\vtconv(c', c, \theta)=\mathbf{1}_{\theta=\text{Id}}(\theta)$ with the indicator function always returning 1 at the group identity element $\text{Id} \in G$ and 0 otherwise. Pointwise group convolutions generalise commonly used $1\times 1$ convolutional filters to (sub)groups. The simplification does not impact the equivariance property (to both rotation and translation) in any way.

Pointwise group convolutions are easier to implement because the summation over the (sub)group filter domain can be omitted, reducing memory and computational cost in the forward-pass by a factor proportional to the (sub)group size $|G|$. In \cref{tab:separable-convolutions}, we compare full, separable and pointwise factorisations of the rotation subgroup in a roto-translation equivariant CNN. We also note an interesting equivalence between applying group convolutions that are pointwise in specific (sub)groups and invariance (to specific subgroups) that is achieved by pooling outputs after applying the same network to augmented inputs on the group. This connects invariance obtained by pooling after stacked layer-wise equivariant layers with DeepSets \citep{zaheer2017deep} and the set-up used in most invariance learning literature \citep{benton2020learning,van2021learning,immer2022invariance}. The equivalence only holds for strict equivariance. Pointwise convolutions have the benefit that they can be relaxed using our method and allow place-coded features related to the group to rate-coded features, on a per-layer basis.

\begin{table}[H]
    \centering
    \resizebox{\linewidth}{!}{
    \begin{tabular}{ l l | r c | c c }
    \toprule
    & & & & Rotated CIFAR-10 & \\
    Roto-translation $(Z^2 \rtimes p4)$-CNN & $\vtconv(c', c, x, y, \theta)$ & Parameters & ($\alpha$) & Test accuracy (\%) & \\
    \midrule
    Full, channel matched & $\vtconv(c', c, x, y, \theta)$ & 1025380 & (10) & 73.70 & \\
    Separable, channel matched & $\vtconv_1(c', c, x, y)\vtconv_2(c', c, \theta)$ & 214550 & (10) & 61.31 & \\
    Depthwise Separable, channel matched & $\vtconv_1(x, y)\vtconv_2(\theta)\vtconv_3(c', c)$ & 139735 & (10) & 51.63 & \\
    \midrule
    Full, parameter matched & $\vtconv(c', c, x, y, \theta)$ & 371200 & ( 6) & 69.89 & \\
    Separable, parameter matched & $\vtconv_1(c', c, x, y)\vtconv_2(c', c, \theta)$ & 358463 & (13) & 59.99 & \\
    Depthwise Separable, parameter matched & $\vtconv_1(x, y)\vtconv_2(\theta)\vtconv_3(c', c)$ & 346215 & (16) & 49.77 & \\
    Pointwise, channel+parameter matched & $\vtconv_1(c', c, x, y)\mathbf{1}_{\theta=\text{Id}}(\theta)$ & 338770 & (10) & 66.77 & \\
    \midrule
    Classical CNN, baseline & $\vtconv(c', c, x, y)$ ($\theta$ not convolved) & 338770 & (10) & 57.89 &  \\
    \bottomrule
    \end{tabular}
    }
    \caption{Experimental comparison of roto-translation equivariant convolutional architectures with full, separable and pointwise factorisation of the rotation subgroup. Reported test accuracy.}
    \label{tab:separable-convolutions}
\end{table}

%

\subsection{Sparsified layers}
For sparsified S-FC and S-CONV layers, we use half the number of anchor points $M$ as there would otherwise be parameters in that layer without spatial sparsification. In non-integer, we round upwards.

\subsection{Datasets}
\label{app:datasets}

For MNIST \citep{lecun1998mnist} and CIFAR-10 \citep{krizhevsky2009learning} datasets, we normalise images standardised to zero mean and unit variance, following standard practice. Rotated datasets `Rotated MNIST' and `Rotated CIFAR-10' consists of original datasets but every image rotated by an angle uniformly sampled from along the unit circle. Similarly, images in `Translated MNIST' are randomly translated in x- and y- axes by uniformly sampled pixels in the range [-8, 8].

\subsection{Training details}
\label{app:training-details}

For the CIFAR-10 experiments, we optimise using Adam \citep{kingma2014adam} ($\beta_1{=}0.9, \beta_2{=}0.999$) with a learning rate of 0.01 for model parameters $\vtheta$ and 0.1 for hyperparameters $\veta$, cosine-annealed \citep{loshchilov2016sgdr} to zero. We train with a batch size of 128 for 4000 epochs, and update hyperparameters every 5 epochs after a 10 epoch burn-in. For parameter updates, we use standard data augmentation consisting of horizontal flips and 4-pixel random shifts, and do not augment when calculating the marginal likelihood estimates. For MAP training, we use unit prior variances $\sigma_l{=}\bar{\sigma}_l{=}1$ for all layers $l$. For Laplace training, we use this same setting to initialise our prior but treat them as part of the hyperparameters $\veta$ and optimise them during training. For MNIST experiments, we use the same settings, except that we do not use data augmentation and train for a shorter period of 1000 epochs at a lower initial learning rate of 0.01. All experiments were run on a single NVIDIA~RTX~3090 GPU with 24GB onboard memory.

\subsection{Code}

Code accompanying this paper is available at \url{https://github.com/tychovdo/ella}

\end{document}